\documentclass[runningheads]{llncs}
\usepackage{marvosym} 
\usepackage[T1]{fontenc}
\usepackage{graphicx}
\usepackage{amsmath}
\usepackage{multirow}
\usepackage[table]{xcolor}
\definecolor{mygray}{gray}{.9}
\usepackage{colortbl}  
\usepackage{pifont}
\newcommand{\cmark}{\ding{51}}%
\newcommand{\xmark}{\ding{55}}%
\begin{document}
\title{Medical Image Registration Meets Vision Foundation Model: Prototype Learning and Contour Awareness}
\titlerunning{SAM-assisted Medical Image Registration}
%

\author{
Hao Xu\inst{1}, Tengfei Xue\inst{1}, Jianan Fan\inst{1}, Dongnan Liu\inst{1}, Yuqian Chen\inst{2,4},\\
Fan Zhang\inst{3}, Carl-Fredrik Westin\inst{2,4}, Ron Kikinis\inst{2,4}, Lauren J. O'Donnell\inst{2,4}, Weidong Cai\inst{1}$^($\textsuperscript{\Letter}$^)$
}

\authorrunning{H. Xu et al.}
%
\institute{The University of Sydney, Sydney, Australia  \and
Harvard Medical School, Boston, USA \and 
University of Electronic Science and Technology of China, Chengdu, China \and
Brigham and Women’s Hospital, Boston, USA\\
\email{tom.cai@sydney.edu.au}
}

%
%
%
\maketitle              
\begin{abstract}
Medical image registration is a fundamental task in medical image analysis, aiming to establish spatial correspondences between paired images. However, existing unsupervised deformable registration methods rely solely on intensity-based similarity metrics, lacking explicit anatomical knowledge, which limits their accuracy and robustness. Vision foundation models, such as the Segment Anything Model (SAM), can generate high-quality segmentation masks that provide explicit anatomical structure knowledge, addressing the limitations of traditional methods that depend only on intensity similarity. Based on this, we propose a novel SAM-assisted registration framework incorporating prototype learning and contour awareness. The framework includes: (1) Explicit anatomical information injection, where SAM-generated segmentation masks are used as auxiliary inputs throughout training and testing to ensure the consistency of anatomical information; (2) Prototype learning, which leverages segmentation masks to extract prototype features and aligns prototypes to optimize semantic correspondences between images; and (3) Contour-aware loss, a contour-aware loss is designed that leverages the edges of segmentation masks to improve the model's performance in fine-grained deformation fields.
Extensive experiments demonstrate that the proposed framework significantly outperforms existing methods across multiple datasets, particularly in challenging scenarios with complex anatomical structures and ambiguous boundaries. Our code is available at \url{https://github.com/HaoXu0507/IPMI25-SAM-Assisted-Registration}.

\end{abstract}
\section{Introduction}
Medical image registration refers to establishing spatial correspondence between fixed and moving images, ensuring that their anatomical structures align with each other~\cite{zhang2021deep,VoxelMorph,TransMorph,TransMatch,CorrMLP}. 
Existing unsupervised deformable registration methods~\cite{zhang2021deep,VoxelMorph,TransMorph,TransMatch,CorrMLP,review-2013,review-2016} primarily rely on intensity-based similarity metrics, which lack explicit anatomical knowledge and often result in suboptimal performance, especially in complex scenarios. These limitations underline the need for incorporating additional structural information to enhance registration accuracy and robustness.

However, due to the difficulty in obtaining segmentation masks of medical images and their inconsistency with clinical scenarios, registration methods with auxiliary segmentation masks have not gained significant attention.

\begin{figure}[th]
\includegraphics[width=0.99\textwidth]{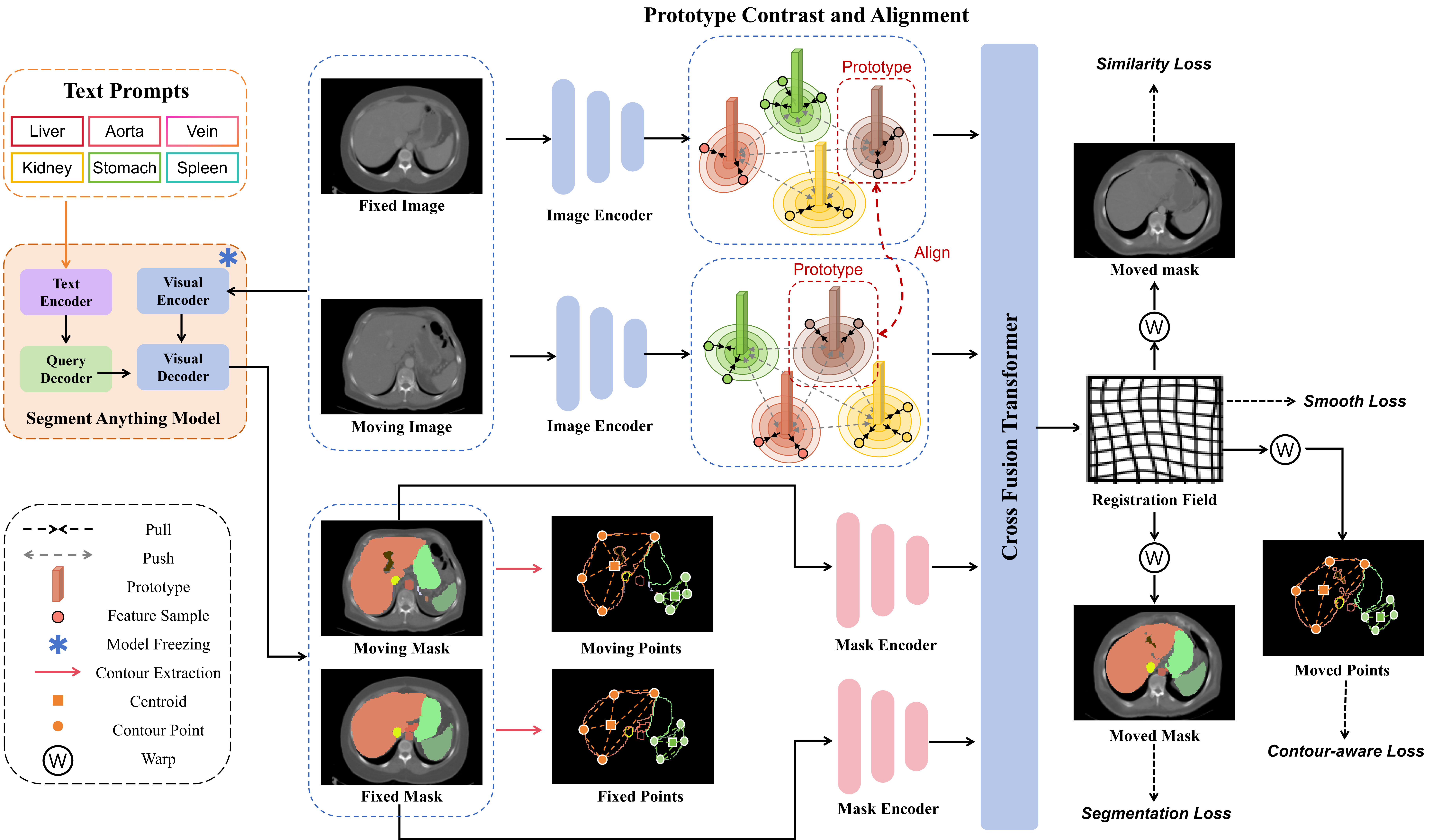}
\caption{The overview of the proposed SAM-assisted medical image registration framework includes the use of the Segment Anything Model (SAM) to generate segmentation masks, optimization of the feature space through prototype contrast and alignment, enhancement of contour alignment via contour-aware loss, and a cross fusion transformer for image and mask feature fusion registration.}  
%
\label{fig.1}
\end{figure}

Segment Anything Model (SAM) and its variants \cite{SAM,MIA-SAM,MedSAM,SAM-review,SAM2D,SAM3D,SAT} gain attention for their powerful segmentation capabilities in both natural and medical images. Current medical-based SAM \cite{MIA-SAM,MedSAM,SAM2D,SAM3D,SAT} can segment whole-body anatomical structures (e.g., different brain regions, heart partitions, abdominal organs) according to the prompts. The common prompts include point, box, mask, and text. Among these, text prompts yield segmentation results comparable in accuracy to the other three types and are more stable. 
Image registration could gain benefits from SAM assistance due to SAM's ability to generate high-quality segmentation masks \cite{SAM-feature-reg}. These masks provide explicit anatomical structure knowledge, which can complement traditional intensity-based methods. 
However, current mask-assisted registration methods have two flaws: i) Since the segmentation mask is inaccessible during the test phase, the segmentation mask is only used to calculate the segmentation loss during the training phase and cannot be integrated into the model, resulting in insufficient utilization of valuable semantic information; ii) Current methods only use segmentation loss, making it difficult to accurately align the contours of organs, especially those with complex shapes.

To address these issues, we propose a novel SAM-assisted registration framework that leverages segmentation masks to explicitly guide the registration process through prototype learning and contour awareness, as shown in Fig. \ref{fig.1}. The proposed framework incorporates three key innovations: i) Explicit anatomical information injection: the segmentation mask generated by SAM is directly used as auxiliary input during the training and testing phases to provide explicit anatomical guidance for the model and ensure the consistency and alignment of anatomical information during the registration process.
ii) Prototype contrast and alignment: the segmentation mask is used to extract the prototype features of each anatomical region, and the features of the same anatomical region are gathered to their corresponding prototypes through contrast learning, thereby improving the consistency of features within the region. At the same time, the prototypes in the image pairs are aligned to ensure that the same anatomical structure across images maintains semantic consistency, and optimizes the registration effect from a global and local level.
iii) Contour-awareness: we design a contour-aware loss function based on the mask contour, focusing on the precise alignment of complex anatomical shapes. This loss function significantly improves the registration accuracy of the model in boundary-sensitive areas.
Extensive experiments show that our model exceeds state-of-the-art methods on the Abdomen CT dataset and ACDC Cardiac MRI dataset.

The main contributions of this work are as follows:
i) We propose a novel SAM-assisted deformable medical image registration framework that utilizes anatomical masks explicitly.
ii) We introduce prototype learning to optimize the semantic consistency of anatomical structures at local and global levels, which enhances the feature disentanglement capability of the model for different anatomical structures.
iii) We design a contour-aware loss to better improve the registration accuracy of the model in the boundary area.
iv) We evaluate the effectiveness of our method on the Abdomen CT dataset and the ACDC Cardiac MRI dataset.

\section{Related Work}

\subsection{Segment Anything Model}
SAM and its variants \cite{CLIPSAM,MIA-SAM,SAM-review,MIA-SAM,SAM3D,SAT,medlsam,zhang2024challenges,Llama} receive widespread attention from the community due to their superior segmentation performance and good generalization. SAM \cite{SAM} consists of an image encoder, a prompt decoder, and a lightweight mask decoder. SAM segments the target according to the given prompts, and the prompt's quality strongly affects SAM's segmentation performance. A good prompt can achieve good results and vice versa. Common prompt methods include point, box, mask, and text. Compared with the first three prompts, the text prompt has the most stable segmentation performance while maintaining the segmentation performance. MedSAM \cite{MedSAM} is the first medical-image-based SAM, which freezes the prompt encoder and fine-tunes the image encoder and mask decoder of the original SAM.
Medical SAM Adapter (Med-SA) \cite{MedSAM} fine-tunes the SAM for medical image segmentation with a learnable adapter layer.
SAM-Med2D \cite{SAM2D} freezes the image encoder of the original SAM and introduces a learnable adapter layer to fine-tune the prompt encoder and update the mask decoder during training. SAM-Med3D \cite{SAM3D} extends SAM-Med2D to a 3D network, further improving its performance.
SAT \cite{SAT} adopts contrastive learning to align medical text and images and uses text prompts to guide the model to segment the corresponding targets. It also establishes a multi-organ and multi-modality dataset for medical image segmentation and achieves good results. However, the provision of text prompts requires some medical background knowledge, which hinders its scope of use.

\subsection{Deformable Medical Image Registration}
Unsupervised deformable registration is to find the spatial correspondence between a pair of images based on their similarity measure.
VoxelMorph \cite{VoxelMorph} is one of the most widely used benchmark registration methods, which is a CNN-based U-Net architecture. Spatial transform is introduced to register the moving image to the moved image according to the field output by the network. 
TransMorph \cite{TransMorph} adopts the paradigm of voxel morph, and the network consists of a vision transformer encoder and a CNN decoder. However, TransMorph simply uses the vision transformer as an encoder, ignoring that the attention mechanism can be approximated as a similarity measure of image patches, which coincides with the image registration. Therefore, TransMatch \cite{TransMatch} adopts the paradigm of voxel morph, proposes to model the spatial correspondence of images using model attention mechanism, and achieves promising results.
The above three methods are single-step registration methods, and it is difficult to achieve good results when the image pairs have large gaps. The multi-step image registration methods split single-step registration into an iterative process. CorrMLP \cite{CorrMLP} is a multi-step registration method, which includes a feature pyramid encoder and a coarse-to-fine MLP registration decoder. By using MLP to model the fine-grained long-distance dependencies of images, it achieves state-of-the-art performance.
However, the unsupervised methods model the spatial correspondence of image pairs based only on image-level metrics, and the performance of the methods is limited due to the lack of medical anatomical information.

Deformable registration with segmentation masks introduces anatomical information, not only relying on the similarity metric of a pair of images but also using the overlap rate of the registered moving mask and the fixed mask as the loss function for model training. The current common paradigm comes from the auxiliary registration module of VoxelMorph. However, this paradigm is limited by the inaccessibility of the segmentation mask during inference, which cannot introduce anatomical information during inference, resulting in limited improvement.

\subsection{Prototype Learning}
\begin{figure*}[th]
\includegraphics[width=\textwidth]{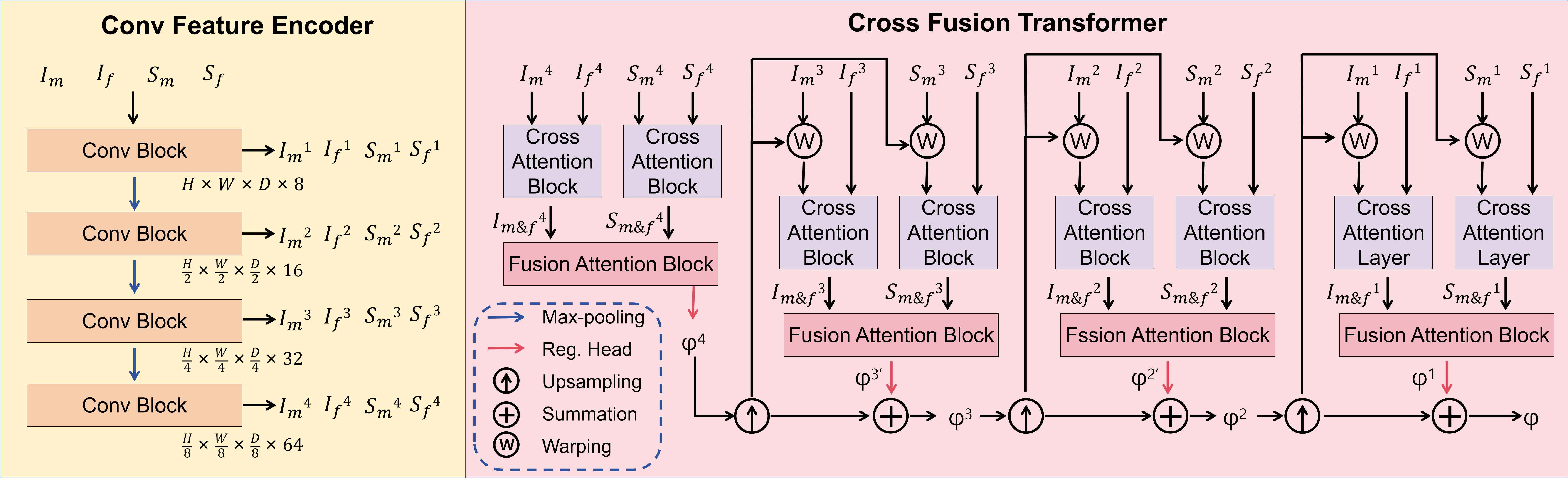}
\caption{The overall architecture of our Encoder and Cross Fusion Decoder.} 
\label{fig.2}
\end{figure*}
Prototype learning \cite{protoSeg,zhang2024prototypical,Prototype-tuning} refers to enhancing the model's ability to represent category semantics by building category prototypes in feature space. In semantic segmentation, prototype-based methods usually leverage masks to aggregate features in feature space, pulling features of the same category closer and pushing features of different categories farther apart to learn the distinguishable category representation space.
Zhou et al. \cite{protoSeg} extract non-learnable pixel prototypes (i.e. cluster centers) from pixels of the same class, and improve the segmentation accuracy by bringing the same class pixels and prototypes closer and pushing different pixels and prototypes farther away. Zhang et at. \cite{zhang2024prototypical} presents prototypical information bottlenecking and disentangling for multimodal survival analysis using pathology and genomics, which extends previous progress thus far on co-attention-based early-based fusion and learning information bottleneck. These methods show the great potential of prototypical learning for optimizing category feature space.

\section{Method}

The goal of medical image registration is to find a spatial transformation that aligns a moving image \( I_m \) with a fixed image \( I_f \) in terms of anatomical structures. Formally, given a pair of images \( I_m \) and \( I_f \), the registration task aims to learn a deformation field \( \phi \) that transform the moving image \( I_m \) to the fixed image \( I_f \):
\[
I_m' = I_m \circ \varphi
\]
where $I_m'$ is the moved image, \( \varphi \) is the spatial transformation function, and \( \circ \) denotes the warp operation.
As illustrated in Fig. \ref{fig.1}, our SAM-assisted framework aims to improve deformable medical image registration by integrating image-level and mask-level features via pairs of images and their corresponding SAM masks.
We are going to sequentially introduce the core components of our framework: SAM Mask Generation, Encoder and Cross Fusion Decoder, Prototype Contrastive Learning and Alignment, Contour-Aware Loss, and the Overall Loss Function.

\subsection{SAM Segmentation Mask Generation}
We utilize the SAM (Segment Anything Model) with text prompts to generate high-quality segmentation masks, which provide explicit anatomical information for the registration process. Given the input image pair \(I_m\) (moving image) and \(I_f\) (fixed image), SAM produces the corresponding segmentation masks \(S_f\) and \(S_m\) according to text prompts \(T=\{t1,...,tn\}\). These masks serve as auxiliary inputs to guide the feature extraction and semantic alignment.

\subsection{Encoder and Cross Fusion Decoder}
As shown in Fig. \ref{fig.2}, our network consists of a CNN feature encoder and a coarse-to-fine cross fusion decoder.
The encoder receives the input image and segmentation mask pairs and extracts multi-scale features through multi-layer convolution operations:
\begin{equation}
I_m^l, I_f^l, S_m^l, S_f^l = Encoder(I_m, I_f, S_m, S_f),
\end{equation}
where $l$ represents the scale level. The spatial resolution of the feature maps decreases progressively while the channel dimension increases to capture both local and global context information.

In the decoder stage, we use the Cross-Attention Block to calculate the correlation of the image feature embeddings and mask feature embeddings to obtain the registration transformation field of the image level and mask level. Then, the Fusion Attention Block is used to aggregate multi-scale features of the image and mask to gradually optimize the deformation field.
The Cross-Attention block \cite{Swin-transformer} consists of four sub-modules: the Window-based Multi-Head Cross-Attention (W-MCA) module, the Fully Connected Network module (MLP), the Shifted Window Multi-Head Cross-Attention (SW-MCA) module, and another MLP module. The cross-attention mechanism \cite{Transformer} is shown as follows:
\begin{equation}
\begin{aligned}
Cross\mbox{-}Attention(Q, K, V) = Softmax(\frac{QK^T}{\sqrt{d}})V,  
\end{aligned}
\label{equation:7}
\end{equation}
where \textit{Q} is the moving image (mask) feature, and \textit{K} and \textit{V} are the fixed image (mask) features. 

The Fusion Attention Block differs from the Cross Attention Block because it introduces bi-directional feature interactions. In the bi-directional computation, the image features are first used as the query $Q$ and the mask features as the key $K$ and value $V$. Then, the roles are reversed: the mask features are used as the query (\(q\)), and the image features as the key $K$ and value $V$. The results of these two interactions are averaged to achieve bi-directional information fusion between the image and the mask.
Finally, the crossed image feature $I_{m\And f}$ and the crossed mask feature $S_{m\And f}$ are fed into the Fusion Attention Block to obtain the registration domain $\varphi^4$. Then, $\varphi^4$ will be upsampled, with one side used to transform ${I_m}^3$ and ${S_m}^3$, and the other side used to superposition $\varphi ^{{3}' }$ as the current scaled transformation field $\varphi^3$. Similarly, we can obtain $\varphi^2$ and $\varphi^1$, where $\varphi^1$ is the ultimate registration field $\varphi$.

\subsection{Prototype Contrastive Learning and Alignment}
To improve the semantic consistency of anatomical regions, we introduce Prototype Feature Extraction, Contrast, and Alignment. First, to enhance the feature consistency within the same anatomical region, we extract prototype features for each anatomical region using Masked Average Pooling based on segmentation masks and align all features to the prototype of the same category. Second, to improve the separability of different prototype features while maintaining semantic consistency across images, we push away different prototypes and align the same prototype between different images. 

\subsubsection{Prototype Feature Extraction}
We leverage the segmentation mask \( S_m \) to extract prototype features for each anatomical region in the moving image \( I_m \) using Masked Average Pooling:
\begin{equation}
P_m^k = \frac{\sum_{i,j} I_{m(i,j)} \cdot S_{m(i,j)}}{\sum_{i,j} S_{m(i,j)}},
\end{equation}
where \( P_m^k \) is the prototype feature for the \( k \)-th region, \( I_{m(i,j)} \) represents the feature at position \( (i, j) \) in the moving image, and \( S_{m(i,j)} \) is the corresponding mask value.
Similarly, we get \( P_f^k \) for the fixed image \( I_f \) and the corresponding mask \( S_f \).

\subsubsection{Prototype Contrast and Alignment}
A contrastive loss is applied to pull features in \( I_m \) corresponding to the same anatomical structure closer to their prototype while pushing features of different structures apart:
\begin{equation}
\mathcal{L}_{\text{contrast}} = - \log \frac{\exp(\text{sim}(I_{m(i,j)}, P_k))}{\sum_{k'} \exp(\text{sim}(I_{m(i,j)}, P_{k'}))},
\end{equation}
where \( \text{sim}(x, y) \) denotes cosine similarity.
Prototypes from the moving image \( I_m \) and fixed image \( I_f \) are aligned to ensure semantic consistency across the corresponding anatomical regions. 
The alignment loss based on cosine similarity can be defined as:
\begin{equation}
\mathcal{L}_{\text{align}} = \sum_k \left( 1 -\text{sim}(P_f^k, P_m^k)  \right), 
\end{equation}
where \( k \) iterates over all anatomical regions.
We use the sum of $L_{contrast}$ and $L_{align}$ as the prototype loss:
\begin{equation}
L_{prototype} = L_{contrast} + L_{align}. 
\end{equation}
\subsection{Contour-Aware Loss}
One of the key challenges in medical image registration is achieving precise alignment of anatomical contours, especially in regions with complex structures or ambiguous edges \cite{SAT-Morph}. To address this issue, we introduce the Chamfer loss \cite{Charmfer} as the contour-aware loss to optimize contour alignment. Specifically, first, we sample point sets $C_f$ and $C_m$ from the contours of the fixed mask $S_f$ and the moved mask $S_m'$. Second, we calculate the Chamfer loss between $C_f$ and $C_m$. The Chamfer Loss is defined as:
\begin{equation}
\mathcal{L}_{\text{contour}} = \frac{1}{|C_m|} \sum_{i \in C_m} \min_{j \in C_f} \|i - j\|^2 + \frac{1}{|C_f|} \sum_{j \in C_f} \min_{i \in C_m} \|j - i\|^2,
\end{equation}
where $\|i - j\|^2$ denotes the squared Euclidean distance between points $i$ and $j$. By minimizing this loss, the model learns to align the contour of $S_m$ with $S_f$, thereby improving the precision of contour alignment.

\subsection{Overall Loss}
In addition to the above prototype loss and contour loss, the overall loss also includes the following common registration losses: similarity loss, smooth loss, and segmentation loss.
\subsubsection{Similarity Loss}
Similarity loss is to measure the similarity of the fixed image $I_{f}$ and the deformed moving image $I_{m'} = Warp(I_m, \varphi^1)$. We adopt a widely used image similarity loss called local normalized cross-correlation (LNCC) loss \cite{VoxelMorph,Transformer,TransMatch}:

\begin{equation}
\begin{aligned}
&L_{sim}(I_{f},I_{m'})  = \\
&\sum_{P\in \Omega }^{} \frac{ ( \sum_{P_i}^{}(I_{f}(P_i)-\hat{I_{f}}(P))(I_{m'}(P_i)-\hat{I_{m'}}(P)) )^{2} }
{(\sum_{P_i}^{}(I_{f}(P_i)-\hat{I_{f}}(P))^2)(\sum_{P_i}^{}{(I_{m'}(P_i)-\hat{I_{m'}}(P))^2})}.
\end{aligned}
\label{equation:8}
\end{equation}

\subsubsection{Smooth Loss}
We use the spatial gradient of diffusion regularization as the deformation field of smoothing loss:
\begin{equation}
L_{smooth} = \sum_{p \in \Omega}^{}  || \bigtriangledown \varphi(p) ||^{2}, 
\label{equation:10}
\end{equation} 
where \textit{p} denotes the voxel location. 
\subsubsection{Segmentation Loss}
Dice loss is adopted as the segmentation loss:
\begin{equation}
L_{seg} =  1-\frac{2\left | S_{m'}\cap S_{f} \right | }{\left |S_{m'} \right | + \left | S_{f} \right |  },
\label{equation:11}
\end{equation}
where $S_{m'}$ is the deformed moving mask.

Finally, the overall loss is as follows:
\begin{equation}
L = \omega_{1} L_{sim} +  \omega_{2} L_{smooth}+ \omega_{3} L_{seg} +  \omega_{4} L_{prototype} + \omega_{5}L_{contour},
\label{equation:11}
\end{equation}
where $\omega_{1}$, $\omega_{2}$, $\omega_{3}$, $\omega_{4}$, and $\omega_{5}$ are hyperparameters.

\section{Experiments and Results}
\subsection{Datasets}
We evaluate our method on two widely validated datasets of different parts of different modalities, inter-patient Abdomen CT (3D Abdomen Organs) \cite{AbdomenCT}, and ACDC MRI (4D Cardiac) \cite{ACDC}.
The inter-patient Abdomen CT dataset is from Learn2Reg \cite{review-2022}, which contains 30 CT scans. We follow the previous settings \cite{CST-registration,li2023samconvex} and randomly select 20 scans for training and 10 scans for testing, pairing them to generate 190 pairs of training sets and 45 pairs of testing sets. Each scan contains manual annotations of 12 anatomical structures: spleen, right kidney, left kidney, esophagus, liver, stomach, aorta, inferior vena cava, portal and splenic vein, pancreas, left adrenal gland, and right adrenal gland. These anatomical regions are also used as text prompts to guide SAM in generating masks. The voxel resolution of all images is 2 ${mm}^3$, and the spatial resolution is 160$\times$160$\times$160. 

The 4D ACDC Cardiac dataset contains end-systolic (ES) and end-diastolic (ED) frames of cardiac MRI images of 100 patients, including 20 healthy patients, 20 patients with previous myocardial infarction, 20 patients with dilated cardiomyopathy, 20 patients with hypertrophic cardiomyopathy, and 20 patients with abnormal right ventricle. Each image contains three anatomical structures: left ventricular (LV), right ventricular (RV), and myocardium (Myo), which are used as text prompts for SAM. 
We aim to register the ED and ES frames of the same patient. We follow the experimental settings of previous works \cite{DiffuseMorph,FSDiffReg} and randomly divide them into 90 pairs for training and 10 pairs for testing. The voxel resolution of all images is 1.5$\times$1.5$\times$3.15 ${mm}^3$, and the spatial resolution is 160$\times$160$\times$160.

\begin{table}[htb]
\centering

\caption{Quantitative comparisons on the Abdomen CT dataset. $\uparrow$: higher is better, $\downarrow$: lower is better. Bold indicates the highest DSC. Spl, Kid R, Kid L, Eso, Liv, Sto, Aor, IVC, Vei, Pan, Adr R, and Adr L represent Spleen, Kidney Right, Kidney Left, Esophagus, Liver, Stomach, Aorta, Inferior Vena Cava, Portal Vein and Splenic Vein, Pancreas, Adrenal Gland Left, and Adrenal Gland Right, respectively. AVG is the average DSC of all 12 organs.
}

\scalebox{0.72}{

\begin{tabular}{|c|cccccccccccc|c|c|}
\hline
\rowcolor{mygray}
 & \multicolumn{13}{c|}{DSC (\%) $\uparrow$}                                                                                         & \\ 
 \cline{2-14}
\rowcolor{mygray}
\multirow{-2}{*}{Methods}                       & Spl  & Kid R & Kid L & Eso  & Liv  & Sto  & Aor  & IVC  & Vei  & Pan  & Adr R & Adr L & \multicolumn{1}{c|}{AVG}  &                 \multirow{-2}{*}{SDlogJ $\downarrow$}         \\ \hline
Initial                   & 30.5 & 23.0  & 27.2  & 10.9 & 50.2 & 19.2 & 21.9 & 20.1 & 1.9  & 7.1  & 3.7   & 3.8   & 17.1 & -                     \\ \hline\hline
VoxelMorph~\cite{VoxelMorph}               & 61.1 & 55.3  & 55.6  & 30.7 & 70.1 & 31.2 & 44.4 & 44.0 & 16.7 & 19.3 & 18.3  & 14.4  & 38.4 $\pm$ 16.6 & 0.143                   \\
TransMorph~\cite{TransMorph}               & 60.3 & 54.3  & 54.0  & 30.1 & 70.7 & 33.5 & 45.4 & 46.1 & 19.3 & 18.5 & 18.2  & 16.3  & 39.0 $\pm$ 16.2 & 0.254                   \\
TransMatch~\cite{TransMatch}               & 65.3 & 58.4  & 56.3  & 33.4 & 72.3 & 36.4 & 49.3 & 54.5 & 21.3 & 20.6 & 19.9  & 18.7  & 42.2 $\pm$ 14.7& 0.101                   \\
CorrMLP~\cite{CorrMLP}                  & 68.2 & 60.1  & 63.7  & 38.4 & 73.4 & 45.6 & 51.2 & 56.9 & 24.7 & 29.1 & 24.6  & 24.0  & 46.7 $\pm$ 13.2 & 0.099                   \\ \hline\hline
\rowcolor{mygray}
SAM Masks ~\cite{SAT}                & 92.8 & 88.1  & 89.5  & 72.2 & 95.5 & 89.4 & 88.4 & 84.5 & 68.5 & 79.7 & 64.2  & 62.6  & 81.2 & -                     \\ \hline
VoxelMorph~\cite{VoxelMorph} + SAM               & 64.2 & 61.1  & 59.6  & 40.7 & 70.3 & 44.4 & 52.1 & 53.5 & 19.2 & 34.1 & 24.8  & 25.9  & 45.8 $\pm$ 12.7 & 0.065                   \\
TransMorph~\cite{TransMorph} + SAM               & 71.0 & 68.7  & 68.6  & 49.2 & 73.1 & 47.4 & 66.3 & 61.1 & 21.6 & 38.2 & 33.9  & 31.7  & 52.6 $\pm$ 12.2 & 0.088                   \\
TransMatch~\cite{TransMatch} + SAM               & 74.0 & 70.4  & 69.6  & \textbf{70.2} & 74.3 & 54.3 & 67.1 & 67.9 & 32.8 & 52.9 & \textbf{46.3 } & 47.1  & 60.5 $\pm$ 11.4 & 0.095 \\
CorrMLP~\cite{CorrMLP}  + SAM                 & 77.1 & 72.8  & 75.7  & 55.1 & 77.7 & 57.2 &\textbf{76.1} & 73.1 & 35.1 & 50.0   & 44.6  & 41.8  & 61.3 $\pm$ 11.3 & 0.103                   \\ \hline
\rowcolor{mygray}
Ours             & \textbf{83.9} & \textbf{79.3}  & \textbf{76.0}  & 55.2 & \textbf{83.8}& \textbf{82.5} & 74.5 &\textbf{73.2} & \textbf{36.9} & \textbf{56.4} & 44.8  & \textbf{48.5}  & \textbf{66.3 $\pm$ 10.6} & 0.091                   \\ \hline

\end{tabular}
}

\label{table:1}
\end{table}

\begin{table}[tbh]
\centering
\caption{Quantitative comparisons on the ACDC Cardiac MRI dataset. $\uparrow$: higher is better, and $\downarrow$: lower is better. Bold indicates the highest DSC. LV, Myo, and RV represent the left ventricular, myocardium, and right ventricular, respectively. AVG is the average DSC of LV, Myo, and RV. }

\scalebox{0.95}{
\begin{tabular}{|c|ccc|c|c|}
\hline
\rowcolor{mygray}
           & \multicolumn{4}{c|}{DSC (\%) $\uparrow$}  &  \\  \cline{2-5}
  \rowcolor{mygray}
  \multirow{-2}{*}{Methods}            & LV   & Myo  & RV   & AVG &    \multirow{-2}{*}{SDlogJ $\downarrow$}    \\ \hline
Initial                                &  58.1& 35.8 & 74.5 &  56.1   & -                     \\ \hline   \hline 
VoxelMorph~\cite{VoxelMorph}  & 83.2 & 60.1 & 80.5 & 74.6 $\pm$ 7.1 & 0.041 \\
TransMorph~\cite{TransMorph}  & 82.5 & 58.8 &80.4  & 73.9  $\pm$ 7.4  & 0.031 \\
TransMatch~\cite{TransMatch} & 82.3 & 58.6 & 82.1 & 74.4 $\pm$ 6.9 & 0.037 \\
CorrMLP~\cite{CorrMLP}    & 82.8 & 72.9 & 83.2 & 79.7  $\pm$ 4.6 & 0.054  \\ \hline
\hline
\rowcolor{mygray}
SAM Masks~\cite{SAT} & 89.2 & 79.2 & 76.2 &  81.6   &-  \\\hline 
VoxelMorph~\cite{VoxelMorph} + SAM   & 86.2 &57.9  & 82.0 &  75.4 $\pm$ 6.3 & 0.077 \\
TransMorph~\cite{TransMorph} + SAM   & 86.0 & 60.7 & 81.3 &  76.0 $\pm$ 5.5  &0.026  \\
TransMatch~\cite{TransMatch} + SAM  & 87.6 & 61.8 & 82.8 & 77.5 $\pm$  5.1& 0.073    \\
CorrMLP~\cite{CorrMLP} + SAM     & 83.4 &  74.2&\textbf{84.2}  & 80.6 $\pm$ 4.2 & 0.047   \\ \hline
\rowcolor{mygray}
Ours                & \textbf{91.9} & \textbf{77.6} & 83.9 & \textbf{84.6 $\pm$ 3.7}    & 0.049   \\ \hline

\end{tabular}
}

\label{table:2}
\end{table}

\begin{figure*}[h]
\centering
\includegraphics[width=\textwidth]{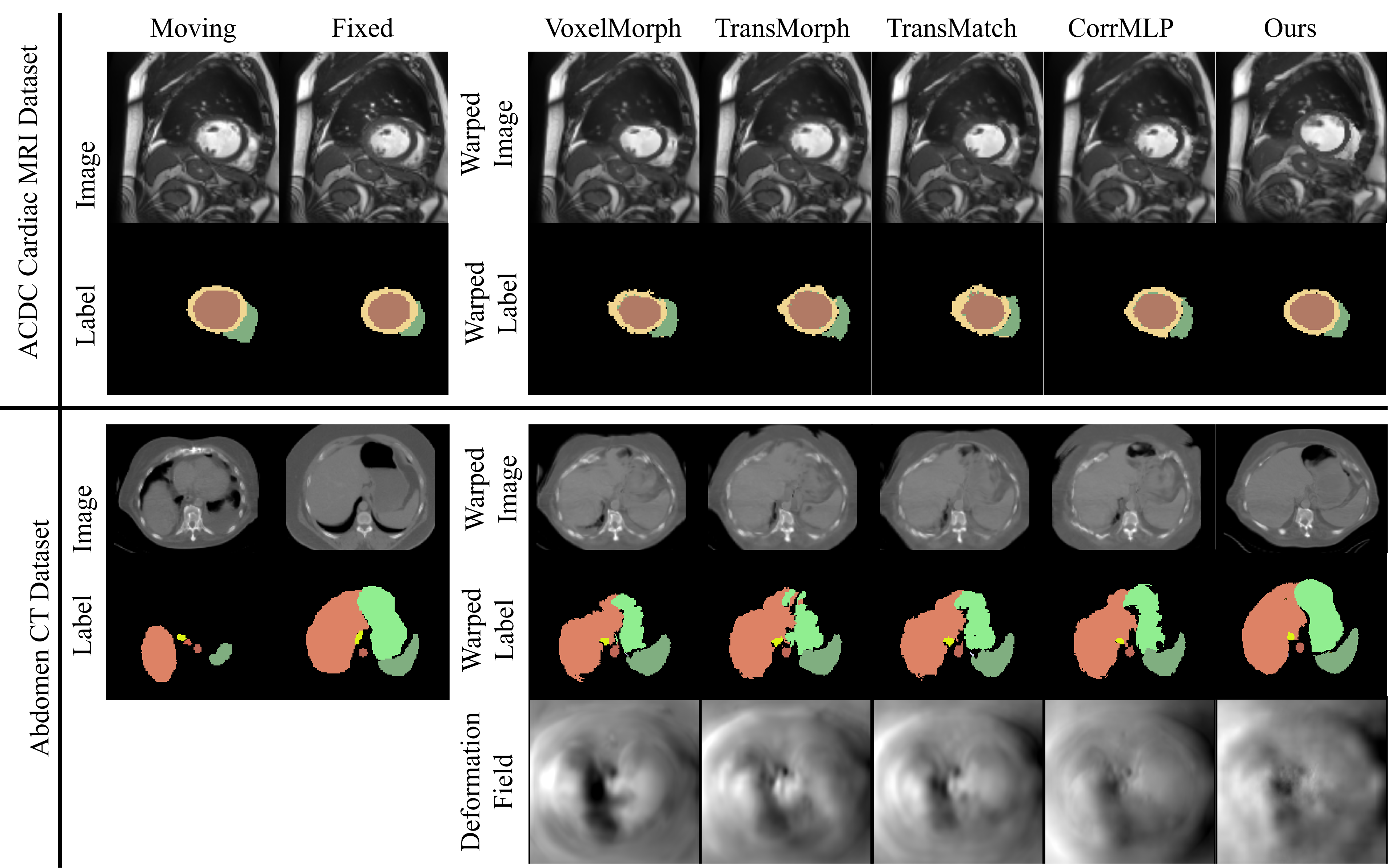}
\caption{Visualization of registration results for our proposed method and compared methods on ACDC Cardiac MRI dataset and Abdomen CT dataset. The comparison methods are all weakly supervised registration methods utilizing SAM masks.} 
\label{fig.3}
\end{figure*}

\begin{table}[h]
\centering
\caption{Analysis of voxel-based contrastive loss on Abdomen CT and ACDC Cardiac MRI datasets. $\uparrow$: higher is better, $\downarrow$: lower is better.}
\begin{tabular}{|c|c c |c|c|}
\hline
\rowcolor{mygray}

Dataset &  $L_{prototype}$  & $L_{contour}$  & DSC (\%) $\uparrow$   & SDlogJ $\downarrow$ \\ \hline
\multirow{3}{*}{Abdomen} & \xmark  & \xmark                              & 62.8  & 0.089   \\
 & \cmark & \xmark                       & 63.3  & 0.086   \\
 \rowcolor{mygray}

& \cmark & \cmark                       & 66.3  & 0.091   \\ \hline
\multirow{3}{*}{ACDC} & \xmark &  \xmark                           & 83.0  & 0.053   \\
                           & \cmark & \xmark                     & 83.5  & 0.045   \\ 
\rowcolor{mygray}
                           & \cmark & \cmark                     & 84.6  & 0.049   \\ \hline
\end{tabular}

\label{table:3}
\end{table}

\subsection{Implementation Details}
We utilize the SAT-Nano model as our SAM model \cite{SAT}. Our experiments are implemented using PyTorch \cite{pytorch} 1.10 on a Nvidia RTX-4090 GPU with 24 GB Memory. Our experiments and all comparison experiments are trained with a batch size of 1 for 500 epochs, using the Adam optimizer with a learning rate of 0.0001. The hyperparameters $\omega_{1}$, $\omega_{2}$, $\omega_{3}$, $\omega_{4}$, and $\omega_{5}$ are 1, 4, 1, 1, and 0.1, respectively. 

\subsection{Evaluate Metrics}
We adopt two widely used evaluation metrics, dice score (DSC) and standard deviation of the Jacobian determinant (SDlogJ). The DSC metric is used to evaluate the overlap of important anatomical regions before and after registration, and SDlogJ is used to evaluate the smoothness of the registration transform field generated by the registration model. The larger the DSC, the higher the overlap and the better the registration result. The smaller the SDlogJ, the smoother the registration transform domain and the better the registration result. 

\subsection{Comparison with the SOTA Methods}

\subsubsection{Comparison Methods}
We compare our method with state-of-the-art deformable image registration methods, including the unsupervised methods VoxelMoprh \cite{VoxelMorph}, TransMoph \cite{TransMorph}, TransMatch \cite{TransMatch}, and CorrMLP \cite{CorrMLP}, and their weakly supervised methods using the SAM segmentation masks. For fair comparison, all weakly supervised methods use the same SAM masks.

\subsubsection{Abdomen CT Inter-Patients Registration}

As shown in Table \ref{table:1}, on the Abdomen CT dataset, our method achieves superior performance of 12 organs compared to all SOTA methods under both unsupervised and weakly supervised setups. Specifically, our method achieves an average DSC of 66.3\%, outperforming the SOTA method CorrMLP by a significant margin while maintaining competitive SDlogJ.
The reason is that other SOTA methods discard the SAM mask in the inference phase, resulting in poor performance due to the lack of explicit information of the anatomical structure. In contrast, our method leverages SAM masks in both training and inference phases to ensure that the model can fully utilize anatomical information, thereby improving the accuracy of image registration.

\subsubsection{ACDC Cardiac MRI Registration of Diastolic and Systolic Phase}
Table \ref{table:2} compares our methods against several SOTA methods on the ACDC dataset. As seen, \textbf{\textit{Ours}} outperforms the CorrMLP by 4.0$\%$. Moreover, \textbf{\textit{Ours} }achieves the highest DSC in both LV and Myo, verifying the effectiveness of our method.

\subsection{Ablation Study}
To evaluate the effectiveness of the proposed prototype loss and contour loss, we conduct an ablation study on the Abdomen CT and ACDC MRI dataset, as shown in Table \ref{table:3}. Specifically, using only the $L_{prototype}$, the DSC of our method increases by 0.5\% from 62.8\% to 63.3\% on the Abdomen dataset and by 0.5\% from 83.0\% to 83.5\% on the ACDC dataset, proving that the prototype loss can improve the semantic consistency to improve the registration accuracy. 
Furthermore, with $L_{prototype}$ and $L_{contour}$, the DSC of our method further increases by 3.0\% up to 66.3\% on the Abdomen dataset and by 1.1\% up to 84.6\% on the ACDC dataset, evaluating that $L_{contour}$ can enhance the contour alignment ability of the model and thus improve the registration performance.
The ablation study shows that $L_{prototype}$ and $L_{contour}$ significantly enhance both registration accuracy and contour alignment while maintaining smooth deformations.

\subsection{Visualization}
Fig. \ref{fig.3} illustrates the qualitative comparison of our method with state-of-the-art (SOTA) registration methods w/ SAM masks on the ACDC MRI and Abdomen CT datasets. 
On the ACDC dataset, our method achieves the most precise alignment of anatomical regions compared to other approaches. The warped labels generated by our method align closely with the fixed labels, particularly for the myocardium (brown) and right ventricle (yellow). In contrast, the warped labels of other methods are misaligned, especially in the myocardial region, indicating their limitations in capturing complex anatomical details. On the Abdomen CT dataset, our method demonstrates superior registration accuracy for challenging anatomical structures such as the liver (orange) and spleen (dark green). The warped labels produced by our method exhibit that our shapes are more complete, and the borders are more rounded compared with other methods. These results validate the effectiveness of our contour loss in enhancing contour alignment and improving the overall quality of the registration.

\section{Conclusion}
In this study, we propose a SAM-assisted medical image registration framework that achieves an effective balance between global semantic consistency and local detail alignment through the integration of explicit anatomical knowledge, prototype learning, and contour-aware loss. The framework first leverages segmentation masks generated by SAM to provide explicit anatomical region annotations, ensuring the consistency of anatomical information during the training and testing phases. Next, prototype learning is employed to optimize semantic consistency across paired images. Additionally, contour Loss is introduced to align the contour points of the fixed and moved masks, further improving boundary alignment precision.
Extensive experiments demonstrate that our model outperforms SOTA methods on the Abdomen CT and ACDC Cardiac MRI datasets by effectively addressing the challenges of complex anatomical structures through explicit anatomical guidance and contour-aware optimization.

\bibliographystyle{splncs04}
\bibliography{mybio}
%




\end{document}